\documentclass[10pt,twocolumn,letterpaper]{article}

\usepackage{iccv}
\usepackage{times}
\usepackage{epsfig}
\usepackage{graphicx}
\usepackage{amsmath}
\usepackage{amssymb}
\usepackage{adjustbox}
\usepackage{lipsum}
\usepackage{subcaption}
\usepackage{multirow}
\usepackage[accsupp]{axessibility} 

\iccvfinalcopy 

\usepackage[pagebackref=true,breaklinks=true,letterpaper=true,colorlinks,bookmarks=false]{hyperref}

\usepackage[capitalize]{cleveref}
\crefname{section}{Sec.}{Secs.}
\Crefname{section}{Section}{Sections}
\Crefname{table}{Table}{Tables}
\crefname{table}{Tab.}{Tabs.}

\def\iccvPaperID{4253} 

\ificcvfinal\fi

\newcommand{\bfsection}[1]{\noindent\textbf{#1.}}

\usepackage{xcolor}
\definecolor{red}{rgb}{0.74,0.08,0.10}
\definecolor{green}{rgb}{0.26,0.49,0.18}
\definecolor{blue}{rgb}{0.22,0.53,0.75}
\newcommand{\name}{CLIPTrans}

\newcommand{\corp}{\mathcal{D}}
\newcommand{\clipimg}{\mathrm{Enc}_{v}^{\mathrm{CLIP}}}
\newcommand{\cliptxt}{\mathrm{Enc}_{t}^{\mathrm{CLIP}}}
\newcommand{\bartenc}{\mathrm{Enc}_{l}^{\mathrm{BART}}}
\newcommand{\bartdec}{\mathrm{Dec}_{l}^{\mathrm{BART}}}
\newcommand{\mapping}{\mathrm{MN}}

\newcommand{\real}{\mathbb{R}}

\begin{document}

\title{{\name}: Transferring Visual Knowledge with Pre-trained Models for Multimodal Machine Translation}

\author{
    Devaansh Gupta\textsuperscript{\rm 1,2,}\thanks{Work done as an intern at Boston College}\\
    {\tt\small guptadm@bc.edu}
    \and
    Siddhant Kharbanda\textsuperscript{\rm 3}\\
    {\tt\small skharbanda@microsoft.com}
    \and
    Jiawei Zhou\textsuperscript{\rm 4}\\
    {\tt\small jzhou02@g.harvard.edu}
    \and
    Wanhua Li\textsuperscript{\rm 4}\\
    {\tt\small wanhua@seas.harvard.edu}
    \and
    Hanspeter Pfister\textsuperscript{\rm 4}\\
    {\tt\small pfister@seas.harvard.edu}
    \and
    Donglai Wei\textsuperscript{\rm 1}\\
    {\tt\small weidf@bc.edu}
    \and
    \textsuperscript{\rm 1}Boston College \ \ \ \ \textsuperscript{\rm 2}BITS Pilani\ \ \ \ \textsuperscript{\rm 3}Microsoft India\ \ \ \ \textsuperscript{\rm 4}Harvard University\\
}
\maketitle
\ificcvfinal\thispagestyle{empty}\fi
\begin{abstract}
    There has been a growing interest in developing multimodal machine translation (MMT) systems that enhance neural machine translation (NMT) with visual knowledge. This problem setup involves using images as auxiliary information during training, and more recently, eliminating their use during inference. Towards this end, previous works face a challenge in training powerful MMT models from scratch due to the scarcity of annotated multilingual vision-language data, especially for low-resource languages. Simultaneously, there has been an influx of multilingual pre-trained models for NMT and multimodal pre-trained models for vision-language tasks, primarily in English, which have shown exceptional generalisation ability. However, these are not directly applicable to MMT since they do not provide aligned multimodal multilingual features for generative tasks. To alleviate this issue, instead of designing complex modules for MMT, we propose {\name}, which simply adapts the independently pre-trained multimodal M-CLIP and the multilingual mBART. In order to align their embedding spaces, mBART is conditioned on the M-CLIP features by a prefix sequence generated through a lightweight mapping network. We train this in a two-stage pipeline which warms up the model with image captioning before the actual translation task. Through experiments, we demonstrate the merits of this framework and consequently push forward the state-of-the-art across standard benchmarks by an average of \textbf{+2.67 BLEU}. The code can be found at \url{www.github.com/devaansh100/CLIPTrans}.
\end{abstract}
\section{Introduction}
Over the decades, Machine Translation (MT) has evolved from being rule-based~\cite{nyberg1992kant}, to more intricate probabilistic models~\cite{marcu2002phrase, ding2005machine, lopez2008statistical, koehn2007moses} and recently to end-to-end deep neural networks~\cite{bahdanau2014neural, cho2014learning, vaswani2017attention, sutskever2014sequence} giving rise to the sub-domain of Neural Machine Translation (NMT). Most recent NMT models largely rely on paired textual data and typically make use of transformer-based encoder-decoder models~\cite{vaswani2017attention,li2022label2label} to set impressive benchmarks~\cite{liu2020very,popel2018cuni}. With advancements in the transformer's ability to encode both images and texts in the same latent space~\cite{su2019vl, li2019visualbert, geng2022multimodal, lu2019vilbert}, there has been a rise in works~\cite{dccn, liu2021gumbel, wu2021good, specia2016shared} leveraging images as auxiliary information to provide visual grounding to the translation task to enhance MT systems, a setting known as Multimodal Machine Translation (MMT). 
\begin{figure}[!t]
    \centering
    \includegraphics[width=\columnwidth]{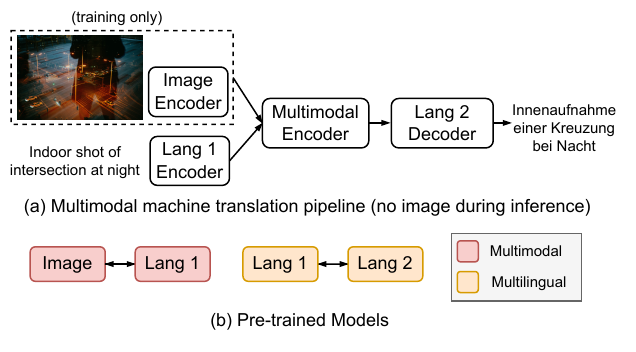}
    \caption{
       (a) Multimodal machine translation (MMT) models are hard to train due to the scarcity of triplet data, especially for low-resource languages.
       (b) Our work aims to leverage existing non-triplet pre-trained models for the MMT task (without image during inference setting).
    } 
\end{figure}
For incorporation of the visual input, previous works have employed specifically engineered encoder-decoder architectures with multimodal attention modules~\cite{dccn, liu2021gumbel, calixto2017doubly, gmnmt, li2021feature, zhou2018visual} that need to be learned from scratch. Consequently, they are forced to balance vision-language alignment with the translation task. 
Furthermore, to reduce the dependence of MMT on images during inference, previous works typically adopt one of two approaches where they either learn a \textit{hallucination} network to generate image features from text~\cite{valhalla, long2020generative}, or use retrieval modules to fetch one or more relevant images~\cite{uvrnmt}. The former requires specially designed losses and difficult optimization while the latter comes with an extra computational cost at test time.

With an increase in popularity of transfer learning methods that make use of task-specific pre-trained unsupervised models, recent NMT works have observed a paradigm shift. 
However, a similar trend has not been witnessed in the MMT domain due to the requirement of data in the form of triplets comprising images and their bilingual captions, which limits transfer learning for three reasons: 
(i) pre-trained models for NMT are only trained on textual data~\cite{conneau2019unsupervised, devlin2018bert, mbart, xue2020mt5} (ii) existing pre-trained models are either multimodal with English as the only language~\cite{li2019visualbert, su2019vl, radford2021learning, tan2019lxmert} or lack decoders for sequence generation~\cite{carlsson-etal-2022-cross, huang2021multilingual} (iii) MMT will require a multilingual multimodal network, which is difficult to train since triplets are expensive to source at the required scale, and existing triplet datasets cannot cover low-resource languages~\cite{shan2022ernie}. 

In this work, we aim to overcome these limitations and simplify the multimodal translation task by employing two independent pre-trained models as aforementioned in (i) and (ii). 
More specifically, we make use of M-CLIP~\cite{carlsson-etal-2022-cross} -- a multilingual variant of the pre-trained multimodal CLIP~\cite{radford2021learning} encoder -- in an optimal training pipeline that tactfully enriches mBART~\cite{liu2020multilingual} -- a pre-trained text-only translation model -- with powerful and well-aligned multimodal features.
CLIP consists of visual and textual encoders that are trained on a large image-captioning dataset using contrastive learning which endows it with generalized, transferable representations for a variety of multimodal tasks~\cite{clipcap, luddecke2022image, ma2022x, li2022grounded}.
When provided with a text input at test time, M-CLIP essentially acts as a \textit{hallucination} network by providing text embeddings pre-aligned with its visual counterpart.
This not only removes the constraint of requiring images during inference but also inherently eliminates the need for hand-engineered architectures with complex training objectives aimed at vision-language alignment~\cite{huang2020unsupervised, suris2020globetrotter}.
Specifically, we employ a mapping network to transfer M-CLIP embeddings as decoder prefix to mBART and train the mBART decoder using a novel two-stage learning pipeline.
In the first stage, we train the mBART decoder for the image-captioning task using a visual-textual decoder prefix sequence computed by a simple, lightweight mapping network from the M-CLIP image encoder. In stage two, the mBART decoder is trained for the translation task, generating decoder prefixes via the M-CLIP text encoder. Interestingly, this mimics the dataset annotation procedure for MMT datasets which first captions an image, then translates the caption while ensuring visual grounding with the image~\cite{specia2016shared, barrault2018findings}. Doing so enables transferring visual representations to the multilingual space, while effectively adapting the mBART attention maps to the newly introduced embeddings.

\bfsection{Contributions}
(1) We present an architecture, \name, that can capitalize on existing pre-trained LMs and multimodal models, thus simplifying the MMT pipeline by eliminating the use of specialized structures and intractable training objectives. 
(2) We propose a novel transfer-learning approach through a two-stage training pipeline wherein the first stage is a shared captioning task and the second is the translation task. We believe we are one of the first works to showcase the merits of using image captioning for adapting pre-trained models for MMT through a thorough analysis and demonstration of quantitative and qualitative results.
(3) We surpass the previous state-of-the-art on MMT across two benchmarks by an average of \textbf{+2.88 BLEU}, and an average of \textbf{+3.64 BLEU} for under-resourced languages, without using images at test time, which further broadens the applicability of our method.
\begin{figure*}[ht!]
    \centering
    \includegraphics[width=\textwidth]{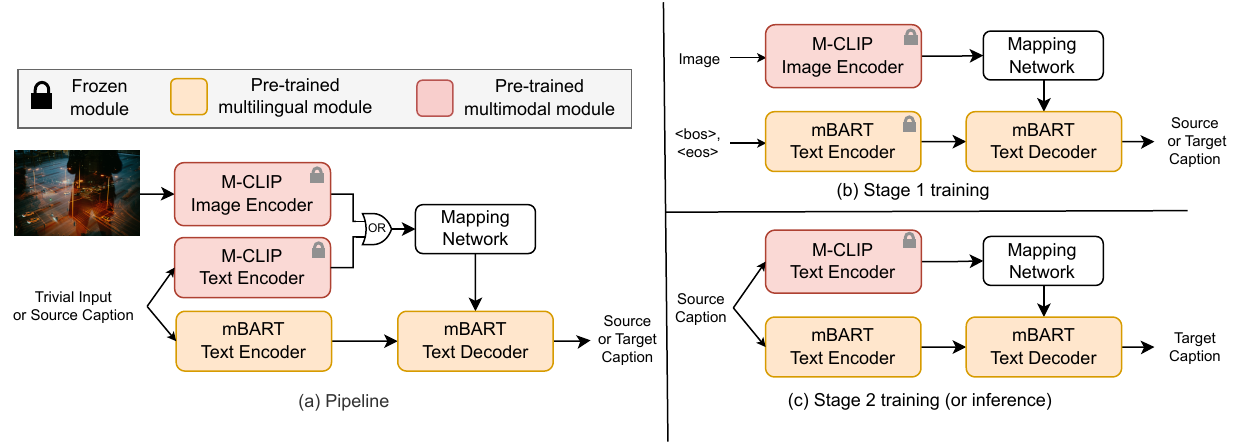}
    \caption{
       \name~framework overview. We show (a) all the modules in {\name} and their wiring to enable transfer learning from pre-trained models for MMT. Along with that, we show the two-stage training pipeline with (b) the image captioning task in the first stage and  (c) the language translation task in the second.
    }  \label{fig:model}
    \vspace{-11pt}
\end{figure*}

\section{Related Work}

\bfsection{Multimodal Machine Translation}
MMT has been examined through various lenses~\cite{specia2016shared, dccn, li2021feature, uvrnmt, valhalla, caglayan2018lium, huang2020unsupervised, calixto2017doubly, liu2021gumbel, gmnmt, zhou2018visual, calixto2017incorporating, yang2020using, sigurdsson2020visual, su2019unsupervised, huang2016attention, peng2022hybridvocab}, with the focus shifting from earlier works on RNN-based encoder-decoder networks to the recently proposed transformer architecture. As discussed earlier, fusion was done through special attention modules. Calixto \etal~\cite{calixto2017doubly} introduces the use of spatial-visual image features through a doubly-attentive attention module and Calixto \etal and Liu \etal~\cite{calixto2017incorporating} further builds upon that to using global visual feature tokens in the source sentence. LIUM-CVC~\cite{caglayan2018lium}, MeMad~\cite{gronroos2018memad} and DCCN~\cite{dccn} use an image-context reweighing of predicted token probabilities while decoding. Gated Fusion~\cite{wu2021good} and UVR-NMT~\cite{uvrnmt} use an image-guided gating mechanism to incorporate image features in decoder cross-attention. In addition to this, UVR-NMT, like RMMT~\cite{wu2021good} also employs a retrieval module to fetch images during inference. Finally, VALHALLA~\cite{valhalla} trains a multimodal encoder and visual hallucination module from scratch for MMT. With respect to using pre-trained models, Kong \& Fan~\cite{bertalibaba} adds a decoding head on top of a BERT model an expensive perform vision-language pre-training, similar to that of VisualBERT~\cite{li2019visualbert}. 
As an alternative to using pre-trained weights, GMNMT~\cite{gmnmt} incorporates a visually grounded multimodal graph built with BERT features into its training data.

\bfsection{Vision-Language Training} Combining vision and language has a long-standing research history. Learning generic cross-modal representations benefits various downstream tasks such as visual grounding~\cite{zhang2020counterfactual}, visual question answering~\cite{goyal2017making}, visual reasoning~\cite{zellers2019recognition}, and visual understanding~\cite{li2022ordinalclip}. Inspired by the success of BERT~\cite{devlin2018bert}, VisualBERT~\cite{li2019visualbert} and VL-BERT~\cite{su2019vl} take both visual and linguistic embedded features as input and train it on the Masked Language Modeling objective. 
VLMO~\cite{bao2021vlmo} proposes a Mixture-of-Modality-Experts Transformer to unify vision-language training models which can process different modalities with a Transformer block. BEiT-3~\cite{wang2022image} further extends it to a multi-way Transformer and attains state-of-the-art results on a broad range of benchmarks. 
While ClipCap~\cite{clipcap} utilizes pre-trained GPT-2 and CLIP to obtain a lightweight image captioning model, 
BLIP~\cite{li2022blip} pre-trains language-image models by bootstrapping the captions.
While these methods show strong generalization ability on various multimodal tasks, they need large vision-language paired datasets and focus on learning multimodal representations. In contrast, we are committed to image-free MMT during inference in a data-constrained setting. 

There have been a plethora of works on transfer learning for machine translation \cite{mbart, raffel2020exploring, xu-etal-2021-bert, mehta2019define, weifinetuned, wang-etal-2022-understanding}. In this work, we propose a training pipeline, along with additional modules, for such models in order to leverage visual information during training to enhance text-only machine translation. More generally, our contribution to the research community can be summarised as a flexible method to enable multilingual generation from multimodal data for MMT, and subsequently to other multilingual seq2seq tasks which can benefit from images.
While this is possible in works like PaLI~\cite{chen2022pali}, PaLM-E~\cite{driess2023palm}, it often cannot be finetuned on downstream data due to closed-source models and/or them being resource-intensive. 

\section{Method}
Let $\corp_v$ denote a vision-based multimodal corpus of image and text pairs $(v, t)$, where $v$ represents an image and $t$ represents the corresponding text.
Let $\corp_l$ denote a language-based multilingual corpus of text and text pairs $(x, y)$, where $x$ represents a sentence in a source language and $y$ represents its translation in a target language.
In MMT, $t$ is either aligned with $x$ or $y$, thus creating a triplet data corpus consisting of $(v, x, y)$ by combining $\corp_v$ and $\corp_l$. 
Our goal is to transfer the knowledge learned with the vision-language corpus $\corp_v$ to augment the task of MT that is conducted on $\corp_l$, with the effective fusion of pre-trained vision-language and language-only models.

\subsection{Preliminaries}
\bfsection{M-CLIP}
Radford \etal~\cite{radford2021learning} proposed the Contrastive Language-Image Pre-training (CLIP) encoders to align vision and language representations in a unified space. It is pre-trained on large-scale image-text paired corpus by matching text descriptions with images. In particular, the model consists of an image encoder $\clipimg$ and a text encoder $\cliptxt$. Given an image-text pair $(v, t)$, the encoded representations $\clipimg(v)$ and $\cliptxt(t)$ are fix-sized vectors that are considered aligned with minimum cosine distance compared with the distances between unpaired texts with the same image.
Although CLIP only works with English, a multilingual CLIP (M-CLIP) that extended the text encoder to work with different languages was also proposed~\cite{carlsson-etal-2022-cross}. 
We rely on the alignment structure of the vision-language representational space of M-CLIP to help transfer the knowledge learned with $\corp_v$ to MT.

\bfsection{mBART}
Pre-trained with sequence-to-sequence denoising objectives, BART (Bidirectional and Auto-Regressive Transformers)~\cite{lewis2020bart} is effective when fine-tuned with various text-to-text generation tasks including MT.
It is composed of a Transformer text encoder $\bartenc$ and a Transformer text decoder $\bartdec$.
Given a source sentence $x=(x_1, x_2, \ldots, x_m)$, BART autoregressively generates the target sentence $y=(y_1, y_2, \ldots, y_n)$ through conditional language modeling
\begin{equation}
\begin{aligned}
    p(y|x) =& \prod_{i=1}^{n} p(y_i | y_{<i}, x) \\
           =& \prod_{i=1}^{n} \bartdec(y_{<i}, \bartenc(x; \theta_e); \theta_d)
\end{aligned}
\end{equation}
where $\theta_e$ and $\theta_d$ are the parameters of the encoder and decoder, respectively, and source sentence $x$ is first encoded by the encoder, and then utilized by the decoder along with the previously generated target $y_{<i}$ for predicting the next token $y_i$.
Different attention mechanisms~\cite{vaswani2017attention} are utilized in the decoder, with the source information $\bartenc(x; \theta_e)$ passed through the cross-attention layers and the prefix information $y_{<i}$ passed through the self-attention layers with autoregressive masks.
For the application of MT,  multilingual BART (mBART)~\cite{liu2020multilingual} that extends BART with pre-training on different languages achieves significant gains when fine-tuned for various MT tasks.

\begin{figure}[ht!]
    \centering
    \includegraphics[width=1.1\columnwidth]{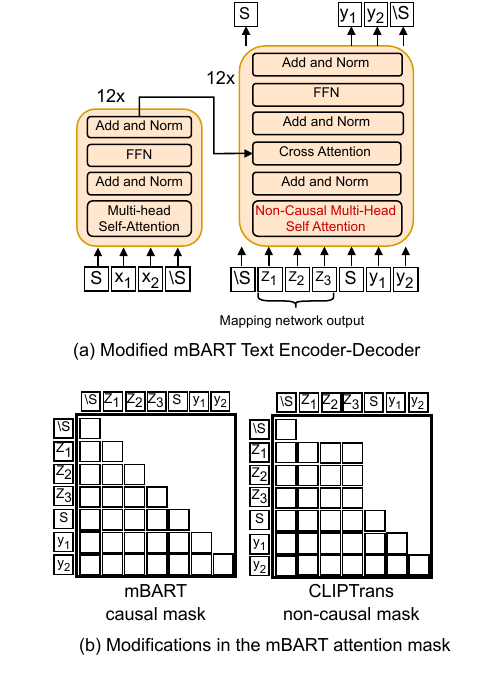}
    \caption{
       (a) Detailed illustration of mBART in {\name}, with modifications in the decoder while training. Note that $x_i$ and $y_i$ are tokens in the source and target language, respectively. $S, \backslash S$ are the special tokens $<bos>, <eos>$. Prefix tokens $z_i$ are concatenated with the shifted output sequence before  decoding. 
       (b) The causal self-attention mask, which masks future tokens to ensure that the next token prediction is done only by attending to the previous ones, is modified to a non-causal one to enable bidirectional information flow amongst the visual context tokens.
    }
    \label{fig:model_details}
    \vspace{-10pt}
\end{figure}

\subsection{Vision and Language Integration}


We aim to integrate vision and language information into a single framework by effectively fusing the multimodal and multilingual pre-trained models, \ie M-CLIP and mBART.
We do so by applying a lightweight mapping network on the M-CLIP encoder representations to produce fixed-length embedding sequences as prefixes prepended to the mBART decoder input.
In particular, we use a simple feedforward neural network for the mapping network, denoted as $\mapping$. Given an encoded M-CLIP representation vector $h\in \real^{d_c} $ either from image $h=\clipimg(v)$ or from text $h=\cliptxt(t)$, it is mapped to a sequence of input embedding vectors for the mBART decoder:
\begin{equation}\label{eq:mapping}
    z = [z_1; z_2; \ldots; z_k] = \mapping(h; \theta_m) \in \real^{k\cdot d_b}
\end{equation}
where $[;]$ denotes vector concatenation, $d_c$ is the visual-textual representation size from M-CLIP, $d_b$ is the embedding size of the mBART decoder, $k$ is the fixed length of the visual prefix embeddings, and $\theta_m$ is the learnable parameters of the mapping network.\footnote{We use a very light feedforward network with no hidden layers, with PReLU activation function on the output.}
Each $z_i\in\real^{d_b}$ for $i=1,\ldots, k$ is serving as a visual-textual prefix token\footnote{This refers to  mapped visual tokens, prepended to textual features.} to be utilized for the text generation with mBART:
\begin{equation}\label{eq:gen}
\begin{aligned}
    p(y|h, x) =& \prod_{i=1}^{n} p(y_i | h, y_{<i}, x) \\
           =& \prod_{i=1}^{n} \bartdec([z, y_{<i}], \bartenc(x; \theta_e); \theta_d)
\end{aligned}
\end{equation}

Moreover, as the visual-textual prefix tokens $z$ are produced all at once, we modify the mBART decoder self-attention mask to be bi-directional for the prefix segment.
An illustration of our model is shown in Fig. \ref{fig:model_details}.

\subsection{Visual Knowledge Transfer Learning}

Based on our integration of pre-trained M-CLIP and mBART, we propose a two-stage learning procedure that utilizes a vision-language corpus $\corp_v$ and a language-only corpus $\corp_l$ separately. The idea is to effectively utilize the internally aligned visual-textual representational structure of M-CLIP for transfer learning between images and texts.

\bfsection{Stage 1: Image-to-Text Captioning}
The first stage is to warm up the mapping network $\mapping$ and the mBART decoder $\bartdec$ to utilize the visual information for text generation.
Given an image-text pair $(v, t)$, we transform the image into visual-textual prefix embeddings based on Eqn.~\ref{eq:mapping} by first passing $v$ into the M-CLIP image encoder, i.e. having $h=\clipimg(v)$, and then applying the mapping network.
We then learn to generate the text $t$ from $v$ based on the autoregressive process modeled by Eqn.~\ref{eq:gen} with mBART, where the target $y=t$, and the source is fixed at $x=(\verb|<bos>|, \verb|<eos>|)$.\footnote{These are two special tokens marking the start and end of a sentence.}
This is essentially an image captioning task, where the image information is encoded in the visual-textual prefix to the mBART decoder, and the caption is generated sequentially after the prefix. The mBART encoder does not provide any information with trivial $x$, which forces the model decoder to rely on the visual-textual prefix information for its generation.
We only update the parameters of the mapping network and the mBART decoder $(\theta_m, \theta_d)$ in this stage, and the M-CLIP and mBART encoders are kept frozen, as shown in Fig.~\ref{fig:model}.

\begin{table*}[!ht]
\centering
\renewcommand\tabcolsep{5pt}
\begin{tabular}{l|c|ccc|ccc|c}
\hline
 \multirow{2}{*}{MMT Model} & \multirow{2}{*}{Inference} & \multicolumn{3}{c|}{EN $\rightarrow$ DE} & \multicolumn{3}{c|}{EN $\rightarrow$ FR} & \multirow{2}{*}{Average} \\ 
 \cline{3-8}
& & Test2016 & Test2017 & MSCOCO & Test2016 & Test2017 & MSCOCO \\ 
\hline
Gumbel-Attention~\cite{liu2021gumbel} & \multirow{6}{*}{L+I} & 39.20 & 31.40 & 26.90 & - & - & - & -6.03\\
CAP-ALL~\cite{li2021feature} & & 39.60 & 33.00 & 27.60 & 60.10 & 52.80 & 44.30 & -4.86\\
GMNMT~\cite{gmnmt} & & 39.80 & 32.20 & 28.70 & 60.90 & 53.90 & - & -4.44\\
DCCN~\cite{dccn} & & 39.70 & 31.00 & 26.70 & 61.20 & 54.30 & 45.40 & -4.71\\
Gated Fusion$^{*}$~\cite{wu2021good} & & 42.00 & 33.60 & 29.00 & 61.70 & 54.80 & 44.90 & -3.43\\
\hline
ImagiT~\cite{long2020generative} & \multirow{7}{*}{L} & 38.50 & 32.10 & 28.70 & 59.70 & 52.40 & 45.30 & -4.98\\
UVR-NMT~\cite{uvrnmt} & & 36.90 & 28.60 & - & 58.30 & 48.70 & - & -7.68\\
VMMT~\cite{calixto-etal-2019-latent} & & 38.40 & 30.10 & 25.50 & - & - & - & -7.19\\
IKD-MMT~\cite{peng-etal-2022-distill} & & 41.28 & 33.83 & 30.17 & 62.53 & 54.84 & - & -5.02\\
RMMT$^{*}$~\cite{wu2021good} & & 41.40 & 32.90 & 30.00 & 62.10 & 54.40 & 44.50 & -3.54\\
VALHALLA~\cite{valhalla} & & 41.90 & 34.00 & 30.30 & 62.30 & 55.10 & 45.70 & -2.88\\
VALHALLA$^{*}$~\cite{valhalla} & & 42.70 & 35.10 & 30.70 & 63.10 & 56.00 & 46.50 & -2.08\\
\textbf{\name~(Ours)} & & \textbf{43.87} & \textbf{37.22} & \textbf{34.49} & \textbf{64.55} & \textbf{57.59} & \textbf{48.83} & \\ 
\hline
\end{tabular}
\caption{Results on the Multi30k dataset. Here we let $^{*}$ represent ensembled models. L+I represents both language and image are used during inference while L means only text is used during inference. \textbf{Bold} represents the highest BLEU score.  We see {\name} outperforms state-of-the-art methods across all settings.}
\label{tab: multi30k}
\end{table*}

\begin{table*}[!htb]
\renewcommand\tabcolsep{11pt}
\centering
\begin{tabular}{l|cc|cc|c}
\hline
\multirow{2}{*}{Model}& \multicolumn{2}{c|}{Under-Resourced} & \multicolumn{2}{c|}{Non-English} & \multirow{2}{*}{Average}\\\cline{2-5}
 & EN $\rightarrow$ RO  & EN $\rightarrow$ AF & DE $\rightarrow$ ES & ES $\rightarrow$ FR\\\hline
RMMT~\cite{wu2021good} & 9.90 & 9.80 & 11.00 & 15.90 & -4.89\\
UVR-NMT~\cite{uvrnmt} & 12.50 & 11.60 & 10.90 & 16.40 & -3.69\\
VALHALLA~\cite{valhalla} & 14.40 & 14.00 & 11.30 & 16.60 & -2.46\\
\name~(Ours) & \textbf{18.34} & \textbf{17.34} & \textbf{13.06} & \textbf{17.41} \\
\hline
\end{tabular}
\caption{Results on the WIT dataset. We observe our method attains the best BLEU scores with a substantial margin.}
\label{tab: wit}
\end{table*}

\bfsection{Stage 2: Text-to-Text Translation}
After stage 1 is done, we further tune the mapping network and the mBART model for the actual translation task relying on the paired textual corpus $\corp_l$ without images.
We swap out the M-CLIP image encoder with the M-CLIP text encoder directly for producing the visual-textual prefix embeddings. 
Specifically, with the translation paired sentences $(x, y)$, we obtain the visual-textual prefix embeddings using Eqn.~(\ref{eq:mapping}) again but with $h=\cliptxt(x)$. We then train the model with translation objectives to generate $y$ from $x$ based on Eqn.~(\ref{eq:gen}).
Note that the source $x$ is passed through both the mBART encoder and the M-CLIP text encoder to be utilized by the decoder for its generation.
The parameters updated in this stage are the mapping network and mBART encoder and decoder, i.e. $(\theta_m, \theta_e, \theta_d)$.
An illustration of this learning stage is shown in Fig.~\ref{fig:model}.

Note that the M-CLIP encoders are kept frozen in both stages. This ensures that its visual-textual representational space does not drift during training. We can utilize this structure to transfer the knowledge learned with visual input (stage 1) to the textual input (stage 2) in the form of the same decoder prefixes, as the visual and textual vectors encoded by M-CLIP are aligned during its pre-training.
As a result, our training objectives are only the text generation cross-entropy loss in both stages,\footnote{No loss is computed on the visual-textual prefix embeddings.} without specially designed auxiliary losses to align the visual and textual information as required by previous approaches~\cite{huang2020unsupervised}.

\subsection{Inference}
Our formulation in Eqn.~\ref{eq:gen} integrates M-CLIP encodings to help MT with the mBART encoder-decoder backbone. The visual-textual representations from M-CLIP allow different application scenarios for MT under our framework. When we have additional input of the image $v$ and $h=\clipimg(v)$, we can achieve vision-based MMT.
For our basic application of text-only MT where we do not have additional image information during inference,
we can simply set $h=\cliptxt(x)$ from the source sentence,
similar to visual hallucination from the text during inference time~\cite{valhalla}.
Decoding can start after the visual-textual prefix computations, either through greedy search or beam search.

\section{Experiments}
\subsection{Experimental Setup}

\begin{table*}[!htp]
\centering
\begin{tabular}{l|cccc|ccc}
\hline
\multirow{3}{*}{Model} & \multicolumn{4}{c|}{Multi30k} &  \multicolumn{3}{c}{WIT} \\ \cline{2-8}
& \multicolumn{4}{c|}{EN $\rightarrow$ DE} & \multirow{2}{*}{ EN $\rightarrow$ RO} & \multirow{2}{*}{EN $\rightarrow$ AF} & \multirow{2}{*}{Average} \\ \cline{2-5} 
& Test2016 & Test2017 & MSCOCO  & Average &  & \\ \hline
\textbf{\name~(Ours)} & 43.87 & 37.22 & 34.49 &  & 18.34 & 17.34 & \\
- Image Captioning & 42.17 & 37.51 & 34.37 & -0.51 & 17.99 & 16.30 & -0.69\\
+ Multilingual Image Captioning & 41.24 & 36.59 & 34.53 & -1.07 & 17.76 & 15.87 & -1.03\\  \hline
\name-reg & 43.40 & 36.44 & 34.67 & -0.36 & 16.69 & 16.21 & -1.39\\
+ Image Captioning & 43.35 & 37.11 & 34.69 & -0.14 & 17.76 & 17.65 & -0.13\\
\hline
mBART & 41.66 & 36.87 & 34.14 & -0.97 & 14.87 & 15.21 & -2.80\\ 
{\name} (M) & 43.40 & 36.44 & 34.67 & -0.36 & 17.27 & 17.31 & -0.55\\ \hline
{\name}-SS & 42.13 & 36.17 & 33.90 & -1.12 & 17.84 & 16.36 & -0.74\\
{\name}-FT & 42.79 & 36.92 & 34.10 & -0.59 & 17.56 & 17.43 & -0.34\\ 
{\name}-CLIP & 42.79 & 37.39 & 34.04 & -1.90 & 18.31 & 17.21 & -0.08\\
\hline
\end{tabular}
\caption{Ablation Results on the Multi30k dataset and WIT dataset.}
\label{tab:ablations}
\end{table*}

\bfsection{Datasets} We demonstrate the effectiveness of our model on two public benchmarks: Multi30k~\cite{elliott2016multi30k} and Wikipedia Image Text (WIT)~\cite{srinivasan2021wit}. 
Multi30k is a widely used MMT benchmark which is a multilingual extension of the Flickr30k dataset that expands EN captions to DE and FR. Evaluation is performed on three standard test splits - Test2016, Test2017, and MSCOCO. 
MSCOCO test split consists of sentences with ambiguous verbs and out-of-domain data points from the COCO Captions dataset, which is considered a generally difficult setting for MMT models~\cite{wu2021good}.
WIT is a multilingual dataset created by extracting image text pairs from Wikipedia in various languages. 
We use this dataset to set new benchmarks on non-English (DE $\rightarrow$ ES, ES $\rightarrow$ FR) and low-resource translations (EN $\rightarrow$ {AF, RO}). Additionally, results on WMT and the EN $\rightarrow$ CS are presented in the supplementary material.

\bfsection{Implementation details} Our models are trained using the previously discussed two-stage training pipeline. Each training stage is trained on 4 A100 GPUs using an AdamW optimizer and Polynomial Decaying Schedule for 15 epochs with a batch size of 256 and a learning rate set to 1e-5. Text decoding is done using beam search with a beam size of 5. All implementations are done in Pytorch using Huggingface Transformers. For the first stage, we pick either the source or target language for captioning depending on their training set alignment in M-CLIP.

\bfsection{Evaluation Metrics} All comparisons are made using BLEU ~\cite{papineni2002bleu}, calculated with SacreBLEU~\cite{sacrebleu}, which is the gold standard for evaluating translation models. Unless otherwise mentioned, we report results using the checkpoint attaining the highest BLEU score on the validation set. We also benchmark our model on the METEOR metric, calculated with the evaluate library\footnote{\url{www.huggingface.co/spaces/evaluate-metric/meteor}}. This can be found in the Supplementary Material.

\subsection{Benchmark Results}
\bfsection{Results on Multi30K} As shown in Table~\ref{tab: multi30k}, our method consistently outperforms all previous state-of-the-art methods and achieves the best BLEU scores across all language-test set splits. 
We compare our architecture with two kinds of methods:  (i) conventional MMT methods that require images during inference and, (ii) methods that do not make use of images during inference. Numbers for comparison are directly quoted from the publication where possible or are obtained using their publicly available codebase.

Specifically, in comparison with the conventional MMT methods that require images during inference, we observe that our method attains +3.43 BLEU improvements on average over the Gated Fusion method~\cite{wu2021good}. These empirical gains validate our model's ability to effectively transfer visual knowledge from M-CLIP models for text-only test time translation.
Next, in comparison with MMT approaches utilizing text-only input during inference, {\name} significantly outperforms UVR-NMT~\cite{uvrnmt} across all metrics without performing multiple image retrieval during inference.
Notably, {\name} outperforms not only the previous state-of-the-art method VALHALLA~\cite{valhalla} by an average of +2.88 BLEU score without training a heavily-engineered hallucination transformer but also its ensemble by a significant margin using only a single instance. 
We attribute these improvements to using pre-trained weights, thus illustrating their effectiveness in MMT.
We observe the highest gains on the difficult MSCOCO test split, which further validates the superiority of our training pipeline at effectively endowing the mBART decoder with visual information.

\begin{figure*}
\centering
\begin{subfigure}{0.33\textwidth}
  \centering
  \includegraphics[width=\linewidth]{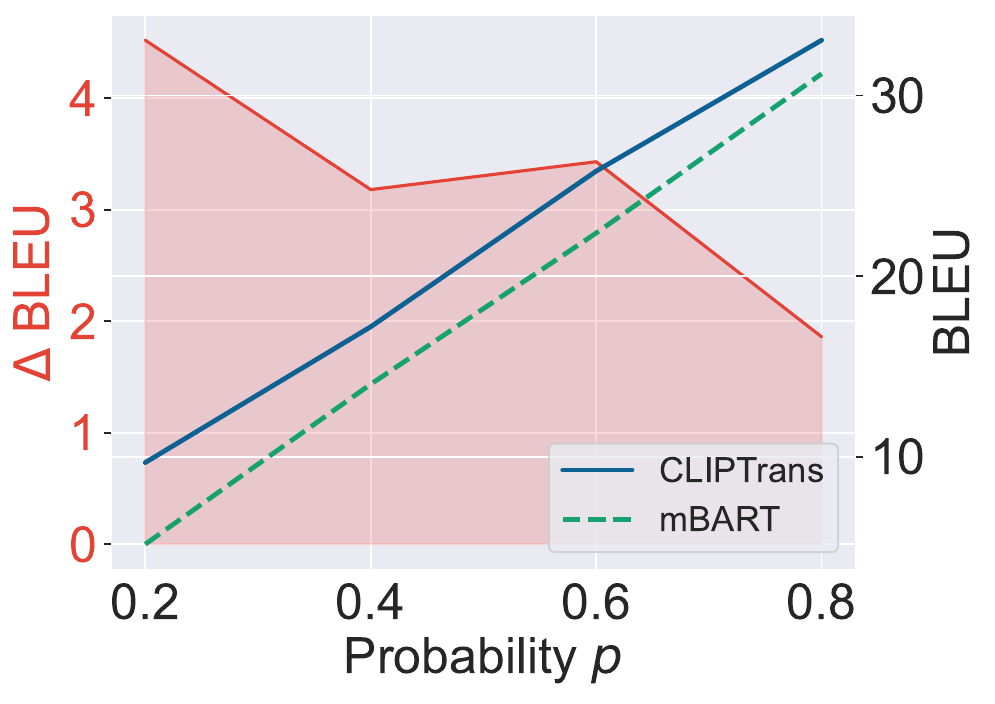}
  \caption{}
  \label{fig:recover_de}
\end{subfigure}
\begin{subfigure}{0.33\textwidth}
  \centering
  \includegraphics[width=\linewidth]{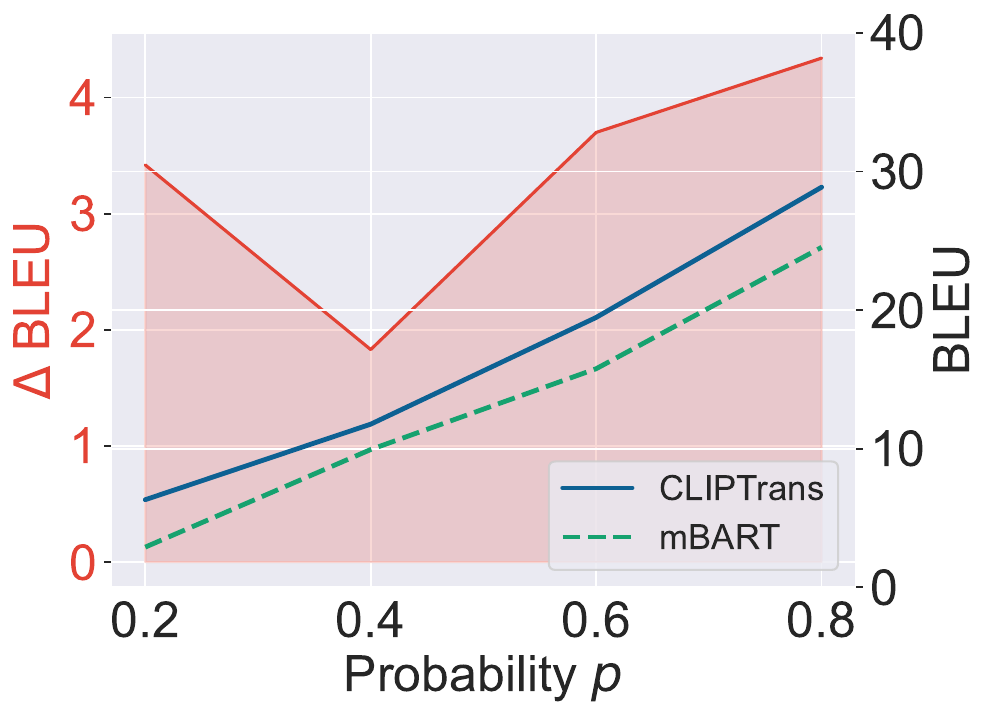}
  \caption{}
  \label{fig:recover_fr}
\end{subfigure}
\begin{subfigure}{0.33\textwidth}
  \centering
  \includegraphics[width=\linewidth]{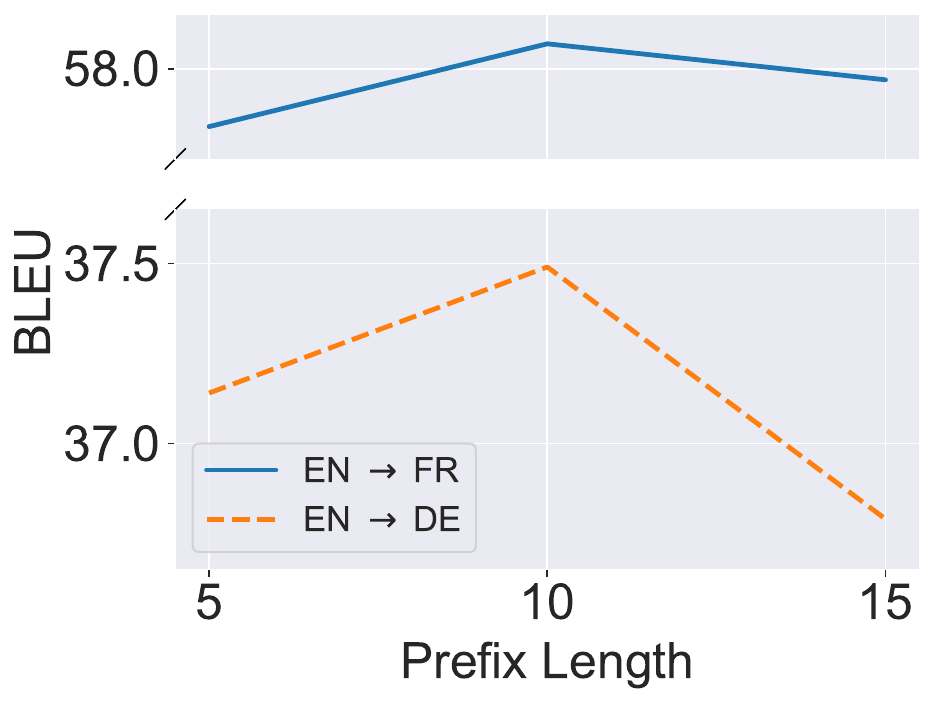}
  \caption{} 
  \label{fig:prefix_length}
\end{subfigure}%
\caption{Evaluation on noisy inputs on {\name} and mBART on the (a) Test2016 and (b) Test2017 split of the Multi30k dataset on EN $\rightarrow$ DE. Recovery is consistently higher than that of an mBART. (c) Sensitivity to the prefix length of different language pairs on the Test2017 split of the Multi30k dataset.} \label{fig:noisy_inputs} 
 \vspace{-15pt}
\end{figure*}

\bfsection{Results on WIT} Tab.~\ref{tab: wit} shows the comparison results on the WIT dataset. We observe that {\name} consistently outperforms existing methods in both under-resourced and non-English settings.  Compared with VALHALLA, our method attains +2.46 BLEU improvements on average, which illustrates the superiority of {\name} over existing methods. Our method shows more significant performance gains on under-resourced settings, where {\name} obtains 3.94 and 3.34 BLEU improvements on the EN $\rightarrow$ RO  and EN $\rightarrow$ AF tasks, respectively.
Relatively smaller gains were seen on non-English benchmarks which can be attributed to two factors (i) there is an English-centric bias in WIT due to which the images are not very well aligned for non-English pairs, as argued in ~\cite{valhalla} and, (ii) imperfect alignment of M-CLIP image-text embeddings for non-English languages since, during training, their representations are derived by machine translating English text to the target language which may introduce inaccuracies.

\subsection{Ablation \& Analysis}
We ablate our training pipeline on both datasets on three language pairs, EN $\rightarrow$ \{DE, RO, AF\}, as shown in Tab.~\ref{tab:ablations}.

\bfsection{mBART} We introduce a new baseline to directly compare the effect of introducing M-CLIP embeddings. Thus, we train a text-only mBART on multilingual captions of each language pair independently. As can be seen in Tab.~\ref{tab:ablations}, mBART is an extremely strong baseline, since it even outperforms ensembles of previous MMT SOTAs on Multi30k and parallels them on WIT. Adding M-CLIP embeddings in {\name} consistently improves upon this baseline, showing the advantage of fusing pre-trained models.

\bfsection{Effect of Image Captioning} In order to understand the benefit of  the image-to-text captioning stage on {\name}, we directly train on translation without the first stage training and report its scores. The performance drops by an average of 0.6 BLEU which shows that captioning is essential for translation in our pipeline and serves as an effective warm-up strategy for the decoder and the mapping network.

\bfsection{Choice of Captioning Language} As mentioned before, we caption on only one language between the source and target, depending on their alignment in M-CLIP. To validate this, we also train our model on both languages (+ multilingual image captioning in Tab.~\ref{tab:ablations}) and notice a performance drop - we conjecture this is because both languages influence the gradients in different directions, thus leading to sub-optimal learning.

\bfsection{{\name} in Traditional MMT Pipelines} Previous MMT approaches~\cite{calixto2017doubly, wu2021good, dccn} used a simple training pipeline where the image and its caption were supplied as inputs and the translated caption as a target. To demonstrate the generalizability of our architecture, we train \name-reg under this setting by passing the image through the M-CLIP Image encoder and the source caption through the mBART encoder. While we notice a drop in performance, more significantly on WIT, we still outperform previous SOTAs. Furthermore, warming up the weights with image captioning brings \name-reg closer to \name, thus validating both, the superiority of our transfer-learning approach and the importance of image captioning for alignment. 

\bfsection{Using Ground-Truth Images in inference} In order to ensure that we perform accurate hallucination during inference, we replace the M-CLIP text encoder with the M-CLIP image encoder and use ground truth images({\name}(M)). We notice a slight drop in performance, as shown in Fig.~\ref{tab:ablations} since this introduces a train-test disparity, as discussed in ~\cite{valhalla}. We further note this disparity when \name-reg is trained with image captioning and tested only with text. 
Thus our approach effectively mitigates this issue without the use of auxiliary losses.

\bfsection{Need for Visual Context} As noted by ~\cite{caglayan2019probing, wu2021good}, images often act as regularizers, especially on the Multi30k dataset, due to the high quality of the paired translation data. They further study the effect of degrading inputs during training and inference, since this would force the model to attend to the images. We believe our image captioning stage enables that and thus demonstrates the ability of {\name} to recover translations, when compared to an mBART trained under the same scenario in Fig.~\ref{fig:noisy_inputs}. For this experiment, we randomly drop tokens from the train and test set with a probability $p$. Furthermore, {\name} uses ground truth images during stage 2 training and inference to study their necessity. While the trends with respect to $p$ are dataset dependent, we consistently see an improvement in {\name} by an average of +3.3 BLEU, even for the low-masking scenario, thus showing the ability of mBART to effectively adapt and utilize the visual context.

\bfsection{Sensitivity to Prefix Length:} We ablate the sensitivity to the prefix length and note that our performance peaks at a length of 10, as shown in Fig.~\ref{fig:prefix_length}. We believe that reducing the prefix length prevents the prefix sequence from being expressive enough while increasing it adds redundancies.

\begin{figure*}[htbp]
    \centering
    \includegraphics[width=0.95\textwidth]{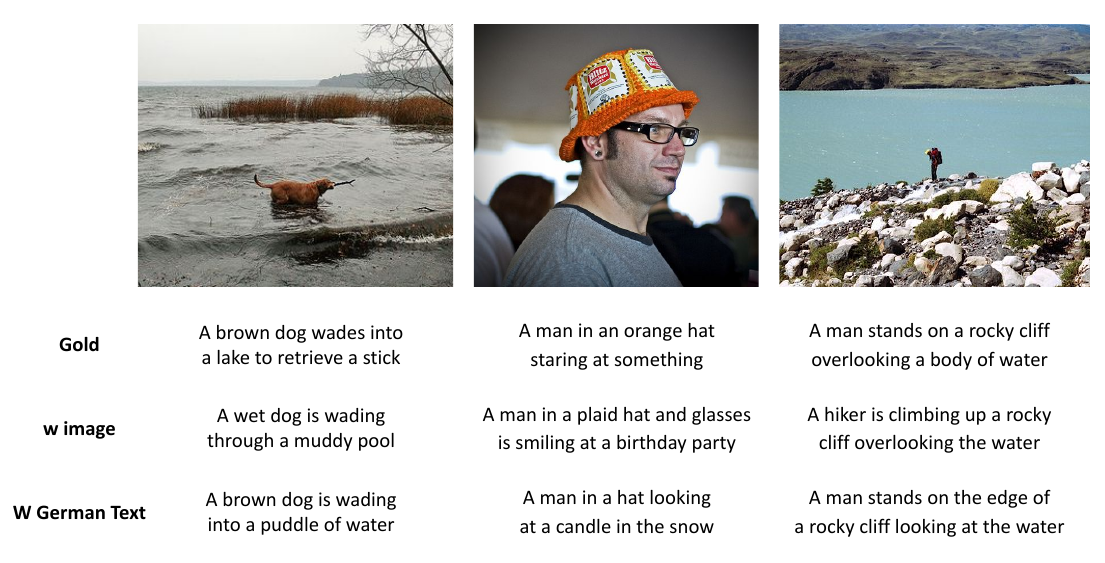}
    \caption{
       Qualitative results of \name~after the captioning stage, on both captioning and zero-shot German to English translation. Data points are from the Test2016 test set of Multi30k. As is visible, CLIP tokens are coherently decoded by the mBART into captions and zero-shot translations.
    } \label{fig: caption}
\end{figure*}

\begin{figure*}[htbp]
    \centering
    \includegraphics[width=0.83\textwidth]{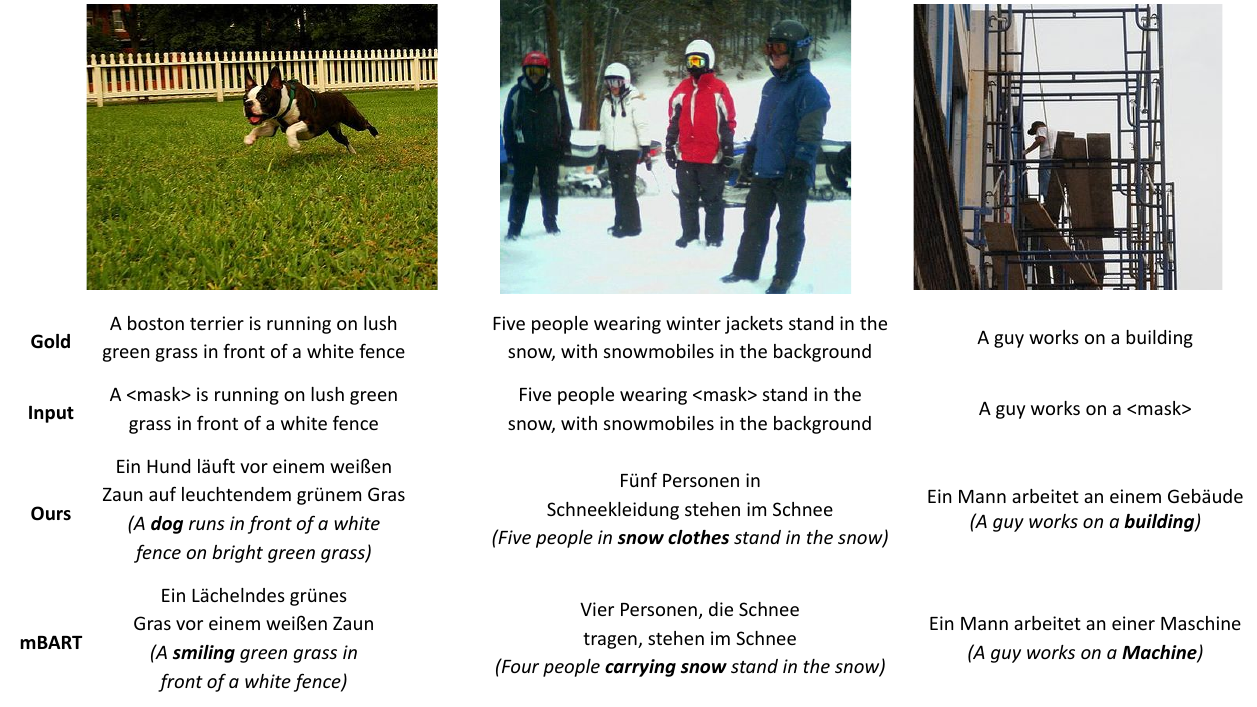}
    \caption{
       Qualitative results of \name~on recovery of visually grounded masked tokens, when compared to an mBART. Data points are from the Test2016 test set of Multi30k. The gold sentence is the ground truth. The \textit{italicized} sentence in the bracket shows the English translation of the German Text obtained via Google Translate, and \textbf{bold} shows the predicted masked word. The image context is effectively utilized and the predicted words are not solely a consequence of the language model, as demonstrated by the mBART translations.
    }  \label{fig: mask_translation}
    \vspace{-3pt}
\end{figure*}

\bfsection{Training Pipelines} We ablate our training pipeline with two variations: (i) single-stage training where we perform stage 1 and stage 2 together. This is done by backpropagating on one data point twice in a batch - once with the image and once with the source text. As shown in Tab.~\ref{tab:ablations}, {\name}-SS gives inferior results than our proposed two-stage pipeline.
(ii) Since we no longer need the CLIP image and text encoder to be aligned after Stage 1, we try finetuning the CLIP text encoder in Stage 2. As can be seen in Tab.~\ref{tab:ablations}, {\name}-FT also attains lower scores. We believe this happens since simultaneous optimization of the mBART encoder, decoder, and CLIP might be difficult.

\bfsection{Choice of image-text encoder} We experiment with different multimodal encoders in our pipeline. Given our setup, having a pre-aligned image-text encoder is imperative. Hence we choose CLIP as presented in \cite{radford2021learning} for this experiment and train {\name}-CLIP in Tab.~\ref{tab:ablations}. Note that M-CLIP uses the same visual encoder as CLIP. Moreover, it facilitates Non-English translations which is not possible with other English-only models. A slight drop in performance shows that M-CLIP's multilingual pre-training creates better language features, even for English\footnote{The sharp average drop on Multi30k in Tab.~\ref{tab:ablations} is largely due to the score on the Test2016 split. We do not see such variations with WIT.}.

\bfsection{Qualitative Results}
In order to further ensure that our results are not solely achieved due to regularisation and that M-CLIP embeddings are not being treated as noise, we show qualitative results that the decoder can actually derive coherent information from them. This is done by using {\name} to decode M-CLIP tokens when no extra information is provided by the mBART encoder. Image captioning results can be seen in Fig.~\ref{fig: caption}. 

We replace the M-CLIP image encoder with its text encoder and evaluate zero-shot translations, by using German text embeddings from M-CLIP to show that captioning knowledge can be transferred to translation due to the inherent structure of M-CLIP. Without extra information from the mBART encoder, we demonstrate in Fig.~\ref{fig: caption} that the decoder can understand a gist of what the translation should be, but does not use the fine-grained context. This context, along with the exact words to be used, is provided when we train with the mBART encoder in the second stage.

Finally, we also show qualitative results for recovery of masked token from the image in Fig.~\ref{fig: mask_translation}. This is done by masking visually grounded phrases in the source text and providing the ground truth image to recover the masked token. We also compare these results with an mBART to ensure that the recovery is a consequence of visual grounding and not a consequence of the language model. We note that while mBART ends up hallucinating phrases, {\name} can recover the phrase accurately.
\section{Conclusion}
This work presents \name, a versatile approach to enable leveraging independent pre-trained models, specifically the multimodal M-CLIP and multilingual mBART, for MMT without using heavily engineered architectures or any external data. Alongside, it presents a two-stage training pipeline wherein the first stage involves an image-to-text captioning task, and the second involves a text-to-text translation task. The efficacy of this schedule and the advantages of transfer learning through image captioning are thoroughly discussed and analyzed. Breaking down the problem as we do further allows us to naturally eliminate the constraint of images during inference without employing complex optimization strategies. 
We not only push the state-of-the-art across multiple datasets, but also set strong text-only baselines with mBART that outperforms previous MMT SOTAs. Given the flexibility of our method, we believe our work could lead future works toward a relaxed MMT setting using unsupervised data. This will enable using existing large-scale datasets during training, thus pushing the domain forward and reducing the reliance on current small-scale datasets.

\section*{Acknowledgements}
This research is supported in part by the NSF award IIS-2239688 and NIH award R01HD104969.
\clearpage
{\small
\bibliographystyle{ieee_fullname}
\bibliography{egbib}
}
\clearpage
\renewcommand{\thepage}{S-\arabic{page}}
\renewcommand{\thesection}{S-\arabic{section}}
\renewcommand{\thetable}{S-\arabic{table}}
\renewcommand{\thefigure}{S-\arabic{figure}}
\setcounter{page}{0}
\setcounter{section}{0}
\setcounter{table}{0}
\setcounter{figure}{0}
\section{Supplementary Material}
\subsection{Language Codes}
The MT language codes mentioned in the paper along with their languages have been shown in Tab.~\ref{tab: lang_codes}.
\begin{table}[!ht]
\centering
\begin{adjustbox}{width=0.7\columnwidth,center}
\begin{tabular}{cc|cc}
\hline
Code & Language & Code & Language\\
\hline
EN & English & ES & Spanish \\
DE & German & RO & Romanian \\
FR & French & AF & Afrikaans \\
CS & Czech & & \\
\hline
\end{tabular}
\end{adjustbox}
\caption{Conventional MT Language codes.}
\label{tab: lang_codes}
\end{table}
\section{Datasets}
\begin{table*}[htb!]
\centering
\begin{adjustbox}{width=0.8\textwidth,center}
\begin{tabular}{c|cc|cccc}
\hline
\multirow{2}{*}{\# samples} & \multicolumn{2}{c|}{Multi30k} & \multicolumn{4}{c}{WIT} \\
\cline{2-7}
 & EN $\rightarrow$ DE & EN $\rightarrow$ FR & EN $\rightarrow$ RO & EN $\rightarrow$ AF & DE $\rightarrow$ ES & ES $\rightarrow$ FR \\
 \hline
Train & 29k & 29k & 40k & 18k & 133k & 122k \\
Validation & 1k & 1k & 5k & 5k & 10k & 10k \\
Test & 2.5k & 2.5k & 1k & 1k & 2k & 2k \\
\hline
\end{tabular}
\end{adjustbox}
\caption{Dataset statistics for Multi30k and WIT}
\end{table*}
\begin{table*}[htb!]
\centering
\begin{adjustbox}{width=0.9\textwidth,center}
\begin{tabular}{c|cccccc}
\hline
& \# Layers & \# Attention Heads & Vocab/Patch Size & Embedding Dim & Feedforward Dim & Projection Dim\\
\hline
mBART & 12 & 16 & 250k & 1024 & 2048 & -\\
XLM-Roberta-Large & 24 & 12 & 250k & 1024 & 4096 & 512\\
ViT-B/32 & 12 & 12 & 32 & 768 & 3072 & 512\\
\hline
\end{tabular}
\end{adjustbox}
\caption{Model statistics for {\name}}
\label{tab: model_stats}
\end{table*}
\subsection{Details}
\bfsection{Multi30k}
Multi30k contains images sourced from the Flickr30k dataset ~\cite{young2014image} with English captions, professionally translated to German and extended to French and Czech. Conventionally, previous MMT methods have reported results only on the German and French splits. The test datasets involve Test2016 and Test2017 which were proposed in their respective years, along with the MSCOCO test set which contains 461 challenging out-of-domain instances from the MSCOCO dataset with ambiguous verbs.

\bfsection{WIT}
WIT is sourced from Wikipedia images and their descriptions in multiple languages. We use this dataset to demonstrate results on low-resource and non-english language splits, specifically on EN $\rightarrow$ \{RO, AF\}, DE $\rightarrow$ ES and ES $\rightarrow$ FR. Apart from this, WIT also contains high-resource splits for EN $\rightarrow$ \{DE, FR, ES\}. These are annotated differently from Multi30k, since the descriptions are independently written for each image, thus inherently introducing noise in the paired translation data and increasing the dependence on images. We use the exact splits as proposed in \cite{valhalla} to ensure uniformity. Note that there can however be some variation in our scores since some images in the training data could not be downloaded. This does not affect the test set due to our text-only setting during inference.

\noindent Whenever needed, we apply preprocessing for both datasets following the input data format of respective pre-trained models.

\subsection{Licences}
All datasets used in this work are publicly available. WIT\footnote{\url{https://github.com/JerryYLi/valhalla-nmt/releases/tag/v0.1-datasets}} \cite{srinivasan2021wit} is available under the CC BY-SA 3.0 license. The license for Multi30k\footnote{\url{ https://github.com/multi30k/dataset}} \cite{elliott2016multi30k} is unknown. Use of images from Flickr30k\footnote{\url{ http://hockenmaier.cs.illinois.edu/ DenotationGraph/}} are subject to Flickr Terms of Use\footnote{\url{https://www.flickr.com/help/terms/}}.
\begin{figure}[!t]
    \centering
    \includegraphics[width=\columnwidth]{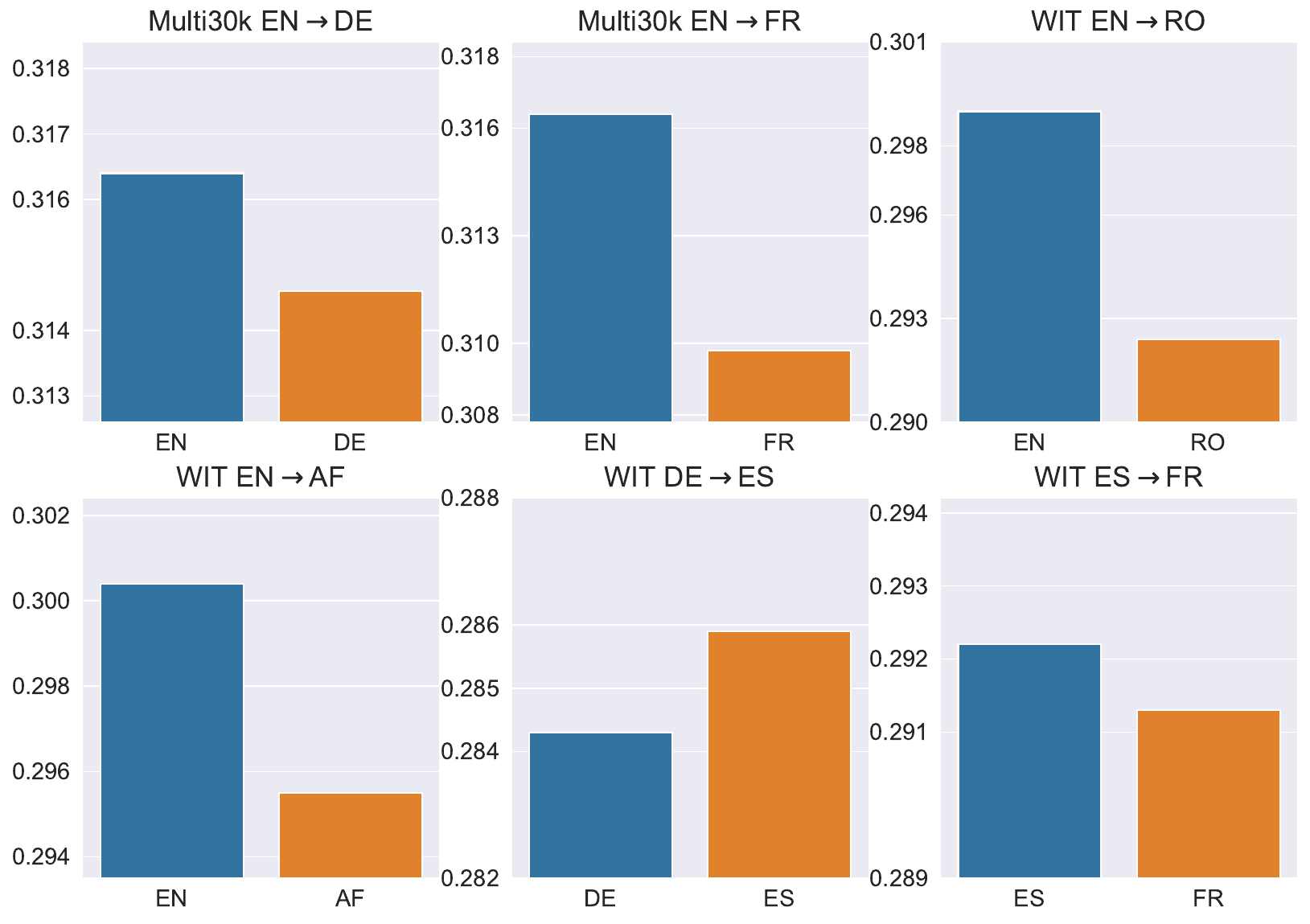}
    \caption{
       Image-caption alignment of all the considered language pairs in their respective training splits. For each split, we perform captioning only on the language with higher similarity.
    }
    \label{fig: caption_lang}
\end{figure}
\begin{table*}[!ht]
\centering
\begin{adjustbox}{width=0.9\textwidth,center}
\begin{tabular}{l|cccc|ccc}
\hline
\multirow{3}{*}{Model} & \multicolumn{4}{c|}{Multi30k} &  \multicolumn{3}{c}{WIT} \\ \cline{2-8}
& \multicolumn{4}{c|}{EN $\rightarrow$ DE} & \multirow{2}{*}{ EN $\rightarrow$ RO} & \multirow{2}{*}{EN $\rightarrow$ AF} & \multirow{2}{*}{Average} \\ \cline{2-5} 
& Test2016 & Test2017 & MSCOCO  & Average &  & \\ \hline
{\name} (Ours) & 43.87 & 37.22 & 34.49 & & 18.34 & 17.34 & \\
\hline
\hline
\multicolumn{8}{l}{Mapping Network Architectures} \\ 
\hline
{\name}-MLP & 41.94 & 35.96 & 33.35 & -1.43 & Unstable & 10.49 & -6.85 \\
{\name}-Enc & 42.29 & 36.75 & 35.41 & -0.37 & 17.86 & 17.54 & -0.13 \\
\hline
\multicolumn{8}{l}{Injection of M-CLIP Embeddings} \\ 
\hline
Before \verb|<eos>| & 43.15 & 38.14 & 34.59 & -0.10 & 17.45 & 16.97 & -0.63 \\
\hline
\end{tabular}
\end{adjustbox}
\caption{Additional Ablations on the Multi30k and WIT dataset}\label{tab: mapping_ablation}
\end{table*}
\subsection{Hyperparameters}
\bfsection{Architectural Details}
We combine two pre-trained models. M-CLIP \cite{carlsson-etal-2022-cross} and mBART \cite{mbart} to develop a multimodal multilingual model. mBART is initialized with its unsupervised pre-trained weights.\footnote{\url{https://huggingface.co/facebook/mbart-large-50}} For M-CLIP we use the model variant consisting of an XLM-Roberta-Large\footnote{\url{https://github.com/FreddeFrallan/Multilingual-CLIP}} text encoder and a CLIP-ViT-B/32 \footnote{\url{https://huggingface.co/openai/clip-vit-base-patch32}} image encoder. The specific configurations of these models is shown in Tab.~\ref{tab: model_stats}.

\bfsection{Choice of Captioning Language}
In the main paper, we demonstrate how captioning on multiple languages harms the performance of the mapping network. Therefore, during the first stage, we perform image captioning using a single language which is chosen on the basis of the image-caption alignment of that language on the training set with M-CLIP. This is calculated by finding the mean cosine similarity of the images and their captions in the M-CLIP encoding space across the training set. A summary of this is shown in Fig.~\ref{fig: caption_lang}.

\subsection{Additional Experiments}
\bfsection{Dependence on Mapping Network Architecture} 
We have chosen the simplest mapping network for our main results, however, we also demonstrate variations of the same by training two additional models with identical hyperparameters -- {\name}-MLP and {\name}-Enc. {\name}-MLP employs fan MLP mapping network with the configuration as Linear$\rightarrow$ReLU$\rightarrow$Linear$\rightarrow$PReLU. {\name}-Enc projects the M-CLIP embedding to the required size, then applies a single transformer layer with two self-attention heads. The results of both are shown in Tab.~\ref{tab: mapping_ablation}. While it may be possible to improve (or stabilize) these results via subsequent hyperparameter tuning, choosing a simple mapping network for {\name}, enables us to set a lower bound on the results.
\begin{table*}[!ht]
\centering
\renewcommand\tabcolsep{5pt}
\begin{tabular}{l|c|ccc|ccc|c}
\hline
 \multirow{2}{*}{MMT Model} & \multirow{2}{*}{Inference} & \multicolumn{3}{c|}{EN $\rightarrow$ DE} & \multicolumn{3}{c|}{EN $\rightarrow$ FR} & \multirow{2}{*}{Average} \\ 
 \cline{3-8}
& & Test2016 & Test2017 & MSCOCO & Test2016 & Test2017 & MSCOCO \\ 
\hline
Gumbel-Attention~\cite{liu2021gumbel} & \multirow{6}{*}{L+I} & 57.80 & 51.20 & 46.00 & - & - & - & -13.97\\
CAP-ALL~\cite{li2021feature} & & 57.50 & 52.20 & 46.40 & 74.30 & 68.60 & 62.60 & -11.40\\
GMNMT~\cite{gmnmt} & & 57.60 & 51.90 & 47.60 & 74.90 & 68.60 & 62.60 & -11.13\\
DCCN~\cite{dccn} & & 56.80 & 49.90 & 45.70 & 76.40 & 70.30 & 65.00 & -10.98\\
Gated Fusion$^{*}$~\cite{wu2021good} & & 67.80 & 61.90 & 56.10 & 81.00 & 76.30 & 70.50 & -2.73\\
\hline
ImagiT~\cite{long2020generative} & \multirow{7}{*}{L} & 55.70 & 52.40 & 48.80 & 74.00 & 68.30 & 65.00 & -10.97\\
RMMT$^{*}$~\cite{wu2021good} & & 68.00 & 61.70 & 56.30 & 81.30 & 76.10 & 70.20 & -2.73\\
VALHALLA~\cite{valhalla} & & 68.80 & 62.50 & 57.00 & 81.40 & 76.40 & 70.90 & -2.17\\
VALHALLA$^{*}$~\cite{valhalla} & & 69.30 & 62.80 & 57.50 & 81.80 & 77.10 & 71.40 & -1.68\\
\textbf{\name~(Ours)} & & \textbf{70.22} & \textbf{65.43} & \textbf{61.26} & \textbf{82.48} & \textbf{77.82} & \textbf{72.78} & \\ 
\hline
\end{tabular}
\caption{METEOR scores on the Multi30k dataset. Here we let $^{*}$ represent ensembled models. L+I represents both language and image are used during inference while L means only text is used during inference. \textbf{Bold} represents the highest score.  We see {\name} outperforms state-of-the-art methods across all settings.}
\label{tab: meteor_multi30k}
\end{table*}
\begin{table*}[!htb]
\renewcommand\tabcolsep{11pt}
\centering
\begin{tabular}{l|cc|cc|c}
\hline
\multirow{2}{*}{Model}& \multicolumn{2}{c|}{Under-Resourced} & \multicolumn{2}{c|}{Non-English} & \multirow{2}{*}{Average}\\\cline{2-5}
 & EN $\rightarrow$ RO  & EN $\rightarrow$ AF & DE $\rightarrow$ ES & ES $\rightarrow$ FR\\\hline
RMMT~\cite{wu2021good} & 23.60 & 29.60 & 33.20 & 36.50 & -4.79\\
UVR-NMT~\cite{uvrnmt} & 28.00 & 32.80 & 32.70 & 37.20 & -2.84\\
VALHALLA~\cite{valhalla} & 30.40 & 34.20 & \textbf{34.30} & 37.50 & -1.41\\ 
\name~(Ours) & \textbf{34.36} & \textbf{35.74} & 34.21 & \textbf{37.73} \\
\hline
\end{tabular}
\caption{METEOR scores on the WIT dataset. We observe our method attains the best scores with a substantial margin.}
\label{tab: meteor_wit}
\end{table*}

\bfsection{Injection of M-CLIP embeddings into mBART}
During pre-training, the first token in the mBART decoder is the  \verb|<eos>| token which has the \verb|<bos>| token as its label. To prevent misalignment with this design choice, we place the prefix sequence after this token. We ablate this and experiment by placing the prefix tokens before it or at the end of the sequence. Subsequently, the decoder self-attention mask is modified. As expected, we notice a slight drop in performance by placing them at the start. Placing at the end causes unstable training for all languages, which can be attributed to the lack of extra self-attention operations undergone by the prefix tokens as compared to placing them at the start, thus preventing them from properly adapting to the mBART.

\bfsection{METEOR} We show the METEOR \cite{denkowski2014meteor} scores on the Multi30k dataset in Tab. ~\ref{tab: meteor_multi30k} and on WIT in Tab. ~\ref{tab: meteor_wit}. Notably, {\name} outperforms all previous SOTAs on METEOR as well.

\bfsection{Additional Results} In order to demonstrate the effectiveness of CLIPTrans for sentences outside the domain of the CLIP pre-training data, we evaluate on WMT2014 for EN→{DE, FR}. Following the undersampled settings in \cite{valhalla}, we take a 100k random subset. Due to the lack of images, we only train stage 2 of {\name}. As can be seen in Tab. ~\ref{tab: wmt_multi}, we outperform the baseline across both languages. 

\noindent For completeness, we also show results in Tab. ~\ref{tab: wmt_multi} the EN $\rightarrow$ CS split of Multi30k, and note that we beat the mBART baseline.
\begin{table}[!htp]
\centering
\begin{adjustbox}{width=\columnwidth,center}
\begin{tabular}{l|cc|cc}
\hline
\multirow{2}{*}{Model} & \multicolumn{2}{c|}{Multi30k(EN $\rightarrow$ CS)} & \multicolumn{2}{c}{WMT} \\ \cline{2-5}
& Test2016 & Test2018 & EN $\rightarrow$ DE & EN $\rightarrow$ FR\\
\hline
mBART & 35.20 & 32.02 & 19.58 & 29.35 \\
{\name} & \textbf{36.05} & \textbf{32.53} & \textbf{21.02} & \textbf{30.34} \\
\hline
\end{tabular}
\end{adjustbox}
\caption{Additional results on WMT and the EN $\rightarrow$ CS split of Multi30k.}\label{tab: wmt_multi}
\vspace{-1.2em}
\end{table}

\subsection{Limitations}
A potential limitation of our method is the computational cost associated with training larger pre-trained models. However, our method is general enough to be replicated on smaller or distilled models as well. Further, in order to take advantages of pre-trained weights, it is limited to the languages used in the pre-training data for M-CLIP and mBART. While this can be counteracted via zero-shot cross-lingual transfer approaches \cite{chen2021zero, tran2020cross}, we leave that for discussion in future works.
\vspace{-0.2cm}

\subsection{Broader Impact}
{\name} can effectively ground images in multiple languages without requiring expensive post-pretraining steps and demonstrates how to effectively leverage exisiting pre-trained models in MMT. Beyond MMT, it can be considered as a generalized approach for developing better multimodal multilingual models using monolingual image captioning data which is of great practical importance. While negative impacts of this are hard to predict, it suffers from the same dataset and societal biases faced by vision and language models. While extensive work is being done to mitigate this, it is beyond the scope of this paper.
\end{document}